\documentclass{article}
\usepackage{ijcai18}
\usepackage{amsmath}
\usepackage{times}
\usepackage{xcolor}
\usepackage{soul}
\usepackage[hidelinks]{hyperref}
\usepackage[small]{caption}
\usepackage{graphicx}
\usepackage{float}
\usepackage[round]{natbib}

\title{An Ensemble Model for Sentiment Analysis of Hindi-English Code-Mixed Data\thanks{This work was presented at $1^{st}$ Workshop on Humanizing AI (HAI) at IJCAI'18 in Stockholm, Sweden.}}
\author{
Madan Gopal Jhanwar$^{\dagger}$, 
Arpita Das\thanks{Both the authors have made equal contributions in this work.}
\\ 
Microsoft India Development Center \\
$\{$majhawar, arpda$\}$@microsoft.com
}

\begin{document}

\maketitle

\begin{abstract}
 In multilingual societies like India, code-mixed social media texts comprise the majority of the Internet. Detecting the sentiment of the code-mixed user opinions plays a crucial role in understanding social, economic and political trends. In this paper, we propose an ensemble of character-trigrams based LSTM model and word-ngrams based Multinomial Naive Bayes (MNB) model to identify the sentiments of Hindi-English (Hi-En) code-mixed data. 
 The ensemble model combines the strengths of rich sequential patterns from the LSTM model and polarity of keywords from the probabilistic ngram model to identify sentiments in sparse and inconsistent code-mixed data. 
 Experiments  on real-life user code-mixed data reveals that our approach yields state-of-the-art results as compared to several baselines and other deep learning based proposed methods. 
\end{abstract}

\section{Introduction}\label{intro}
The rapid growth of opinion sharing on social media has led to an increased interest in sentiment analysis of social media texts. Sentiment Analysis can provide invaluable insights ranging from product reviews to capturing trending topics to designing business models for targeted advertisements. Many organizations today rely heavily on sentiment analysis of social media texts to monitor the performance of their products and take the user feedback into account while upgrading to newer versions.

Social media texts are informal with several linguistic differences. In multilingual societies like India, users generally combine the prominent language, like English, with their native languages. This process of switching texts between two or more languages is referred to as code-mixing. Millions of internet users in India communicate by mixing their regional languages with English which generates enormous amount of code-mixed social media texts. One of such popular combinations is the mixing of Hindi and English, resulting in Hindi-English (Hi-En) code-mixed data. For example, ``yeh gaana bohut super hai''(this song very super is), meaning \emph{``this is a superb song''}, is a Hi-En code-mixed text. 

Apart from several existing challenges such as the presence of multiple entities in the text and sarcasm detection, code-mixing brings with it many other unique challenges. The linguistic complexity of code-mixed content is compounded by the presence of spelling variations, transliteration and non-adherence to formal grammar. The romanized\footnote{
https://en.wikipedia.org/wiki/Romanization} code-mixed data on social media presents inherent challenges like word or phrase contractions (``please'' to ``plz''), and  non-standard spellings (such as ``cooolll'' or ``suppeerrrrr''), etc. Along with diverse sentence constructions, words in Hindi can have multiple variations when written in English which leads to a large amount of sparse and rare tokens. For instance, ``pyaar''(love) can be written as ``peyar'', ``pyar'',  ``piyar'', ``piyaar'', or ``pyaarrrr'', etc.  

Code-mixing is a well-known problem in the field of NLP. Researchers have put in  efforts for language identification, POS tagging and Named Entity Recognition of code-mixed data \citep{bali2014borrowing,chittaranjan2014word,vyas2014pos,kumar2018consonant,sequiera2015overview,solorio2014overview,rao2016cmee}. Over the past years, researchers have established deep neural network based state-of-the-art models for sentiment analysis \citep{socher2013recursive,zheng2018left,ma2018targeted} in English data. For the problem of sentiment analysis of Hi-En code-mixed data, sub-word level representations in LSTM have shown promising results \citep{joshi2016towards,kumar2018consonant}. However, since the code-mixed data is noisy in nature and the available datasets are smaller in size to tune deep learning models, we hypothesize that n-gram based traditional models should be able to assist deep learning based models in improving the overall accuracy of sentiment analysis in code-mixed data.

In this paper, we propose an ensemble model where we combine the outputs of character-trigrams based LSTM model and word ngram based MNB model to predict the sentiment of Hi-En code-mixed texts. While the LSTM model encodes deep sequential patterns in the text, MNB captures low-level word combinations of keywords to compensate for the 
grammatical inconsistencies. Results reveal that our model is able to outperform other traditional machine learning approaches as well as the deep learning models proposed in literature.

The main contribution of the paper are as follows:
\begin{itemize}
    \item  We propose the use of well-established character trigrams as sub-word features in LSTM network that shows comparable performance with other proposed methods. This saves the effort of complicated feature engineering in sparse code-mixed data.
    \item We propose an ensemble of character-trigrams based LSTM model and word-ngrams based MNB model to predict the sentiment of Hi-En code-mixed data.
    \item We evaluated and compared our model with various traditional machine learning classifiers as well as other state-of-the-art techniques. We also present a qualitative analysis of how ngram based MNB model helps overcome some of the shortcomings of LSTM model.
\end{itemize}

Rest of the  paper is organized as follows. We provide an overview of the existing approaches for sentiment analysis of code-mixed data in Section \ref{relwork}. Section \ref{ourapproach} explains various data pre-processing steps taken, the design and training of the ensemble model. In Section \ref{exp},
we explain our experimental setup, describe the performance of proposed system and compare it with
baselines and other methods, proceeded by a discussion of our results. Finally, Section \ref{conc} concludes the paper.

\section{Related Work}\label{relwork}

Information extraction from user-generated code-mixed data is difficult due to its multilingual nature. Language identification tasks have been performed on several code-mixed language pairs \citep{banerjee2014hybrid,mandal2015adaptive,solorio2011analyzing,bali2014borrowing,barman2014code,das2015code}. NLP specific tasks such as POS tagging \citep{solorio2011analyzing,vyas2014pos,jamatia2015part,GuptaTEB17} and NER \citep{rao2016cmee,GuptaEB18} have also been performed on the code-mixed data. Initiatives have been taken by shared task like FIRE-2015\footnote{
http://fire.irsi.res.in/fire/2015/home} to study retrieval of mixed script of Indian languages. However, these proposed solutions do not align with the problem of sentiment analysis in code-mixed data.

Following the current trend, researchers have seen great success in the task of sentiment analysis of English data using deep neural networks. Recurrent Neural Networks (RNN) and its variants have consistently outperformed traditional sentiment analysis state-of-the-art models \citep{socher2011semi,socher2012semantic,socher2013recursive}. \citet{zheng2018left} employed context2target attention based LSTM model to perform targeted sentiment analysis by capturing most important words in left and right context. \citet{ma2018targeted} integrated common sense knowledge into recurrent encoder to form \emph{sentic} LSTM. Due to the availability of large scale labeled English data, the LSTM models are able to capture rich sequential patterns from the data to capture the sentiments. However, the code-mixed data is limited and sparse in nature, making it difficult for the deep learning techniques to learn generic patterns from the data effectively.

In the area of sentiment analysis of Hi-En code-mixed data, very less work has been done so far. A shared task for Sentiment Analysis of Indian Language (Code-Mixed) (SAIL Code-Mixed)\footnote{http://www.dasdipankar.com/SAILCodeMixed.html} on twitter data was organized at ICON-2017\footnote{https://ltrc.iiit.ac.in/icon2017/}. \citet{patra2015shared} summarizes the dataset used, various models submitted by the participants and their results. The best submission for the Hi-En language pair used features like GloVe word embeddings with 300 dimensions and TF-IDF scores of word and character ngrams. They trained an ensemble of linear SVM, Logistic Regression and Random Forests to classify the sentiments.

Among the deep learning approaches,  \citet{joshi2016towards} employed sub-word level representations in LSTM architecture, yielding state-of-the-art result as compared to other traditional machine learning models and word-polarity based models. However, due to the small and very sparse dataset, we believe that the deep learning based techniques cannot capture all the hidden patterns of the data and specifically could not generalize the rare keywords that have impact on sentiment of the sentence. \citet{kumar2018consonant} introduced phonemic sub-word units and used them with a hierarchical Bi-directional LSTM (BiLSTM) model to detect sentiment in Hi-En code-mixed texts. We believe such a complex network with so many weights and hyper-parameters cannot be tuned to its full potential on a small dataset. Therefore, we propose an ensemble model where keyword-based MNB model helps overcome some of the shortcomings of a deep learning based classifier.

\section{Our Approach}\label{ourapproach}

\begin{figure}[!t]
\centering
\caption{Architecture of Ensemble classifier}
\includegraphics[scale=0.45]{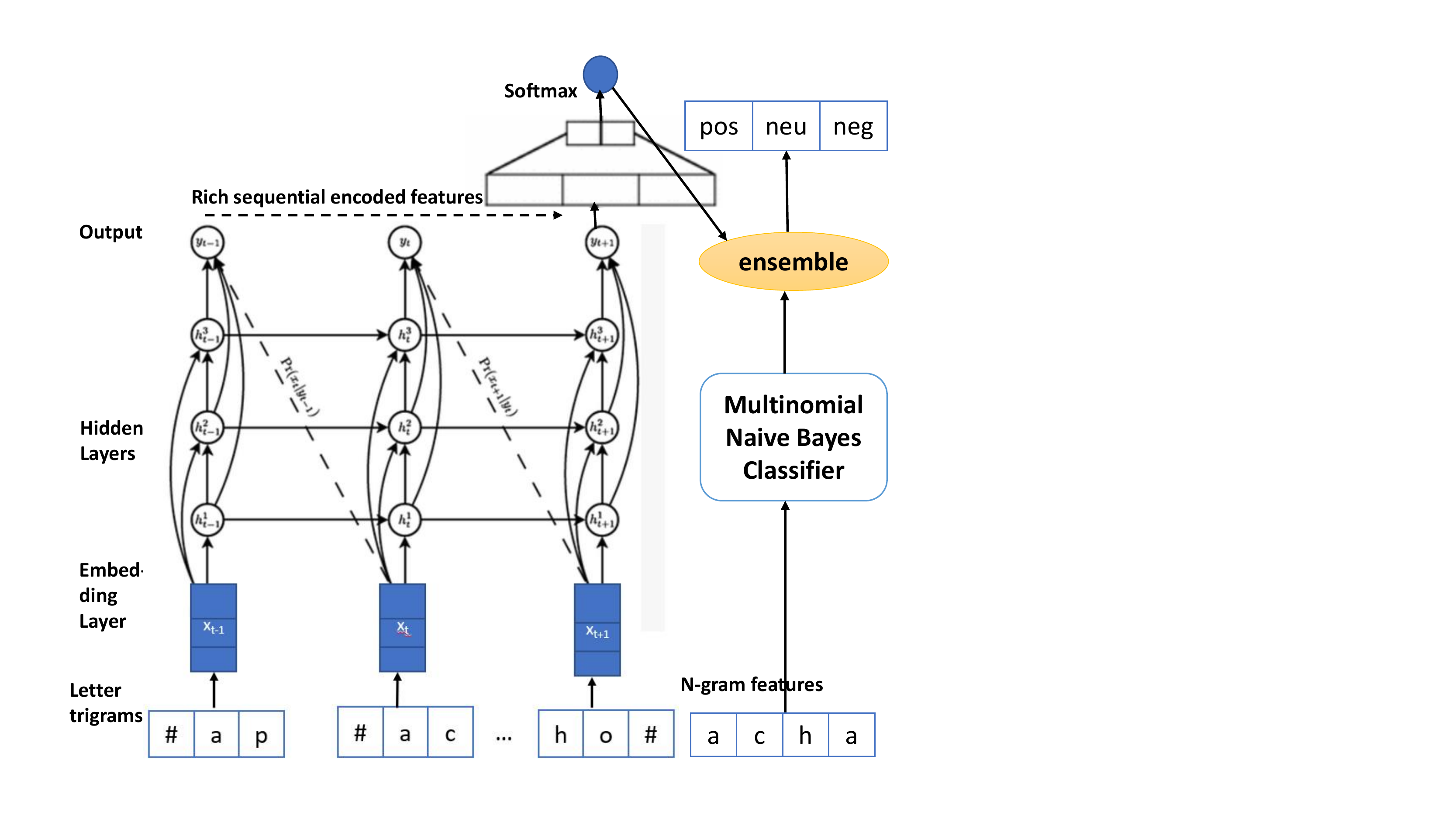}
\label{fig:architecture}
\end{figure}

The architecture of the proposed system is shown in Figure \ref{fig:architecture}. We use a parallel ensemble of two models -- a traditional machine learning model, and an end-to-end deep learning model, to classify a sentence into one of the \emph{positive}, \emph{negative} or \emph{neutral} sentiment classes. 
For the traditional machine learning model, we feed the ngram features of the sentence to a MNB classifier, which outputs the probability of the sentence belonging to each of the classes.
For the deep learning model, the input sentence is fed in the form of character-trigram embedding matrix. The embedding matrix is in turn fed into a LSTM layer which encodes the sequential patterns in the query and outputs a feature representation. This feature representation then passes through a fully-connected (FC) layer, which models the various interactions between these features and outputs the probability of the sentence belonging to each of the three classes. 
We combine the outputs of both of the models to predict the final sentiment of the sentence. We will now explain the details of each of the above mentioned components.

\subsection{Ngram-based Classifier}
After pre-processing the sentence with lower-casing, punctuation and stop-word removal, we generate word-based unigram and bigram features of the sentence and feed them to a MNB classifier. Although known for its simplicity, the Naive Bayes algorithm, introduced by \citet{manning2008text}, is one of the best classifiers in terms of accuracy and computational efficiency \citep{ting2011naive}, and has been widely used for text classification \citep{adi2014classification,zhou2016combining,torunouglu2013wikipedia}.

In this probabilistic learning method, the probability
of a document $d$ being in the sentiment class $c$ is computed as:
\begin{equation}
    P(c|d) \propto P(c) \prod_{1 \leq k \leq n_d} P(t_k|c) 
\end{equation}
where $P(t_k|c)$ is the conditional probability of term $t_k$ occurring in a document of class c and $P(c)$ is the prior probability of a document occurring in class c.
\begin{equation}
    c \in \{positive, negative, neutral\}
\end{equation}

\medskip
In MNB, the best class is the most likely or Maximum Aposteriori (MAP) class.

\begin{equation}
    c_{map} = \arg\max_{c \in C}[log\hat{P}(c) + \sum_{1 \leq k \leq n_d} log\hat{P}(t_k|c)]
\end{equation}

Each conditional parameter
$log \hat{P}(t_k|c)$ represents the weightage of the term $t_k$ for c. The prior $log \hat{P}(c)$ captures the relative frequency of c in C. More frequent classes are more likely to be the correct class than infrequent
classes. The sum of log prior and term weights is a measure of
the evidence for the document to belong to the class and MNB selects the class with the most evidence.

Due to the low availability and high sparsity of the code-mixed training data,  
the problem of zero probability for unknown words is very prominent in this problem setting. To overcome it, we also make use of Laplace smoothing. 

\subsection{Deep Learning based Classifier}
  
We use character-trigrams based Long-Short Term Memory (LSTM) model to capture rich sequential patterns in the sentences to identify their sentiment. We pre-process the sentences by lower-casing them and removing the punctuations and stop-words. In code-mixed language, we often see repetitions of characters. For example, ``yeh''(this)  can be written as ``yehh'', ``yehhhhh'', etc., and ``bohut''(very) can be expressed as ``bohhhhut'', ``booohut'', ``bohuttttt'',etc. We remove the repetitive characters whenever the characters are repeated more than twice ( \emph{``yehhhhh''} $\to$ \emph{``yehh''}). We also append the delimeter ``\#'' to every word (\emph{``main''} $\to$ \emph{``\#main\#''}). After the above mentioned pre-processing steps, we obtain a unique set of 4126 character trigrams. Every token of the input sequence is encoded by 1-of-K character trigrams where K= 4126.

We chose LSTMs, introduced by \citet{hochreiter1997long}, as they
solve the vanishing gradient problem in RNNs at a small computational
cost. They are also able to capture the long-term dependencies
present in a sequential pattern due to their gating mechanisms which
control information flow.

\subsubsection{LSTM Model}

For the 3-class sentiment classification of the code mixed data, we designed a LSTM based classifier with the following details.
\begin{enumerate}
    \item   \textbf{Input Features} : Each token is represented
as a bag-of-character-trigrams vector. We allowed a maximum sequence of 100 character trigram features and applied truncation and padding in case of excess and deficit tokens respectively. For every token, we fed a 128 length embedding matrix to the LSTM unit.

This feature is a fair representation of the sparse code-mixed data as it helps to solve out-of-vocabulary issues and
removes the influence of the word stems, diverse variations and contractions that arises during conversion of Hindi to romanized code-mixed data.

\item \textbf{Output} : The output of the end state of the final LSTM layer is connected to a Fully Connected (FC) layer which models the interactions between
these features and the classes . A softmax activation function is used
to produce correctly normalized probability values.

\item \textbf{Loss function} : We train the parameters of the classifier with an objective of maximizing their
predication accuracy given the target labels in the training set or minimizing the cross entropy error across the set. If $t$ is the true label and $o$ is the output of the network, the cross entropy (CE)
loss function is calculated as follows:

\begin{equation}
    CE(t,o) = −(tlog(o) + (1 − t)log(1-o))
\end{equation}

\end{enumerate}
The optimal hyper-parameter configuration of the classifier set is shown in Table \ref{table:configlstm}.
\begin{table}[!t]
\begin{center}
\begin{tabular}{l|l} 
\hline
\bf{Hyperparameter} & \bf{Value}  \\
\hline
Batch Size & 32\\
Max length & 100 \\
Character Embedding & 128\\
LSTM cells & 64 \\
Learning rate & 0.01 \\
Optimizer & Adagrad \\
 \hline
\end{tabular}
\end{center}
\caption{Hyperparameters of LSTM classifier }
\label{table:configlstm}
\end{table}

\section{Experiments and Results}\label{exp}
In this section, we will give a brief introduction of the dataset used and discuss in detail the various experiments we carried out and their quantitative results.

\subsection{Dataset}
\citet{joshi2016towards} released a dataset for sentiment analysis of Hi-En code-mixed data. The dataset contains user comments from public \emph{Facebook} pages of \emph{Salman Khan}, a Bollywood actor, and \emph{Narendra Modi}, the Prime Minister of India at the time. The dataset contains 3879 sentences, split into 15\% \emph{negative}, 50\% \emph{neutral} and 35\% \emph{positive} classes.

For the purpose of experimentation, we divided the data into three sets -- train set, development set and test set, in the ratio of 70\%, 10\% and 20\% respectively. The train set is used to train the models, development set is used to tune the model parameters and test set is used to evaluate the model performances.

\subsection{Experiments}
As proposed in \citet{pang2008opinion} and \citet{wang2012baselines}, Support Vector Machines (SVM) and MNB have performed well for the task of sentiment analysis on English movie review and customer review datasets. Therefore, for the task of sentiment classification using ngram-based features of Hi-En code-mixed user comments, we experimented with word unigram and bigram features, and evaluated SVM and MNB classifiers.

Among the deep learning methods, \citet{joshi2016towards} proposed sub-word level representations in LSTM to analyze the sentiment of Hi-En code-mixed data and achieved state-of-the-art accuracy. They also implemented the model proposed by \citet{sharma2015text}, where the authors use token sentiment polarity to obtain the final polarity of the sentence, and reported their accuracy on the same dataset. Recently, \citet{kumar2018consonant} proposed using consonant-vowel sequences as phonemic subword units and used a hierarchical deep learning model to predict the sentiment of Hi-En code-mixed data. Since their model is not publicly available, we implemented their system to the best of our knowledge and capabilities for performance comparison.

\begin{figure}[!t][H]
\centering
\caption{Comparison of LSTM vs BiLSTM model loss and error during training phase }
\includegraphics[scale=0.5]{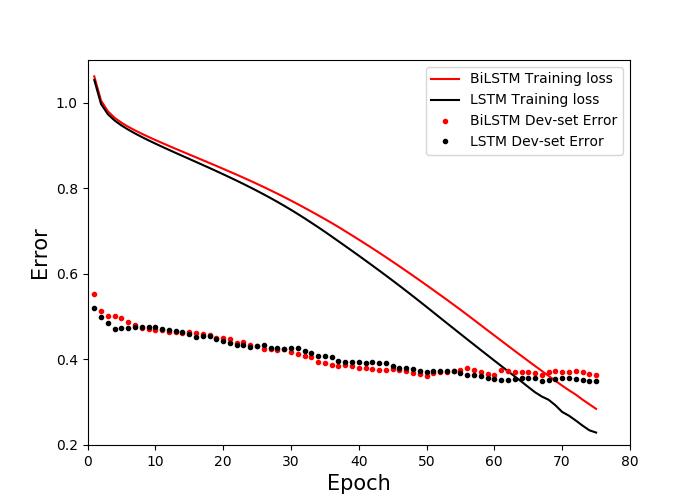}
\label{fig:lstmvsbilstm}
\end{figure}

In this paper, we propose the use of well-established character trigrams as subword features for the task of sentiment classification in Hi-En code-mixed dataset. We evaluated both LSTMs and BiLSTMs on top of character trigram embedding to predict the sentiment of a sentence. As seen in Figure \ref{fig:lstmvsbilstm}, LSTM model is able to reduce the training loss at a better rate than BiLSTM while generalizing on the Dev set as well. This behavior could be owed to the smaller dataset size where LSTM model has lesser number of weights and hyper-parameters to learn.

We further created an ensemble of ngram-based and deep learning classifiers. The word-bigram based MNB classifier and the char-trigram based LSTM model both output the probability of input sentence belonging to all the three classes. We experimented to combine their output probabilities in two possible ways:
\begin{itemize}
    \item We take average of the linear combination of probability outputs of both the models, weighted by their accuracy on the development set, for each class and pick the class with the highest combined probability.
    \item We multiply the output probability of both the models for each class and pick the one with highest combined probability. 
\end{itemize}

The comparison of performance of both the methods on Test set is shown in Table \ref{table:ensembleComparison}.

\begin{table}[!t][H]
\begin{center}
\begin{tabular}{l|l} 
\hline
\bf{Method} & \bf{Accuracy}  \\
\hline
Weighted-Linear Combination & 69.1\\
Multiplication & \textbf{70.8}\\
\hline
\end{tabular}
\end{center}
\caption{Accuracy comparison for combining the outputs n-gram based MNB and char-trigram based LSTM in the model ensemble }
\label{table:ensembleComparison}
\end{table}

Table \ref{table:accSummary} summarizes the quantitative performances of various classifiers discussed above. It can be observed that our Ensemble model outperforms other traditional and deep learning based models on a small Hi-En code-mixed data.

\begin{table*}[!t][h]
\begin{center}
\begin{tabular}{l|l|l|l|l} 
\hline
\bf{Model} & \bf{Accuracy} & \textbf{Precision} & \textbf{Recall} & \textbf{F1-Score}  \\
\hline
SVM (Unigrams) & 61.7 & 0.579 & 0.551 & 0.565 \\
SVM (Unigrams+Bigrams) & 64.1 & 0.609 & 0.537 & 0.566 \\
MNB (Unigrams) & 64.5 & 0.748 & 0.485 & 0.588 \\
MNB (Unigrams+Bigrams) & 66.1 & 0.698 & 0.540 & 0.609 \\
\hline
SentiWordNet \citep{sharma2015text} & 51.15 & - & - & 0.252 \\
\hline
Char-trigram based LSTM & 65.2 & 0.610 & 0.563 & 0.586 \\
Vowel-Consonant based \citep{kumar2018consonant} & 62.8 & 0.652 & 0.522 & 0.580 \\
Sub-word composition based \citep{joshi2016towards} & 69.2 & 0.684 & 0.623 & 0.652 \\
\hline
Ensemble (proposed) & \textbf{70.8}  & 0.718 & 0.612 & \textbf{0.661}\\
\hline
\end{tabular}
\end{center}
\caption{Quantitative comparison of various models proposed for the task of sentiment analysis of Hi-En code-mixed data }
\label{table:accSummary}
\end{table*}

We also perform qualitative analysis to prove the importance of the ensemble approach. In Table \ref{table:qualitative}, we show a comparative study of the performance of the LSTM and MNB models on real-life Hi-En code mixed data. We tried to justify the observations in the \emph{Comments} column of the Table. We observed that LSTM was performing better for sentences with longer length due to its ability to capture sequential information. For the examples that mostly contain rare keywords like ``fadu'' (meaning awesome in English) or slangs, ngram based MNB model performed better than LSTM. Due to lesser occurrences of these keywords in the training data, LSTM is not able to generalize on their impact on sentiment of the sentence, however, ngram based MNB is successfully able to capture it. These observations justify our decision to combine both the models to obtain a better performing system.

\begin{table*}[!t]
\begin{center}
\begin{tabular}{l|l|l|l|l|l} 
\hline
\bf{Example} & \bf{MNB} & \textbf{LSTM} & \textbf{Ensemble} & \textbf{Label} & \textbf{Comments}  \\
\hline
agar mushalman ko challange &&&&&LSTM model is able\\
de rahe ho to bro movi m bhar &	Neu & Neg & Neg & Neg& to capture rich patterns\\
de jholi wali kavali kyo aad ki &&&&&in long sequences.\\
\hline
welcome back mr pm ab kitne &&&&\\
din rahenge india me  ye paisa &&&&\\
kisika baap ka nehi hai hamara &	Neu&Neg&Neg&Neg&Same as above.\\
hai hamare paison se bidesh ja &&&&\\
kar ghumte ho saram nehi aati &&&&\\
\hline
&&&&& MNB captures the keyword\\bhai fadu hai&	Pos	& Neu	& Neu	& Pos &``fadu'', while LSTM could\\ &&&&&not generalize it, due to \\&&&&&its less occurences in data.\\
\hline
abe c****** kabhi namaz bhi &Neg&Neu&Neg&Neg&MNB catches the slang\\
pad liya kar&&&&& keyword as negative. \\

\hline
\end{tabular}
\end{center}
\caption{Qualitative analysis to compare the performance of LSTM and MNB model. The column \emph{Label} represents the gold label sentiment of the example (Positive [Pos], Negative [Neg] or Neutral [Neu]) and the columns \emph{MNB}, \emph{LSTM} and \emph{Ensemble} represent the labels predicted by the MNB, LSTM and Ensemble models respectively. The English translations of the examples in order are -- 1) If you are challenging Muslims, brother, then why did you add the Qawwali ``Bhar de jholi'' in your movie? 2) Welcome back, Mr. PM. How many days will you stay in India now? You go abroad with our money, are you not ashamed? 3) Brother, you are extraordinary!  4) Stupid, pray at least!}
\label{table:qualitative}
\end{table*}
\section{Conclusion}\label{conc}
With the increase in popularity and impact of social media texts, it becomes extremely important to analyze their sentiments to have a understanding of the society. In this paper, we perform sentiment analysis of the sparse and inconsistent Hi-En code-mixed data. We point out the shortcomings of deep learning models on a small multilingual code-mixed data. Further, we propose an ensemble model of n-gram based probabilistic model (MNB) and char-trigram based deep learning model(LSTM) to identify sentiment in code-mixed data. We justify our hypothesis with quantitative and qualitative analysis.

In future, we would like to extend our work to several other language pairs of code-mixed data. It would be interesting to utilize the rich features of individual languages to help identifying sentiments in their code-mixed version. 
\bibliographystyle{named}
\bibliography{ijcai18}

\end{document}